\documentclass{article}

\usepackage{microtype}
\usepackage{graphicx}
\usepackage{booktabs} 
\usepackage{subcaption}

\usepackage{hyperref}

\usepackage[accepted]{icml2021}

\icmltitlerunning{What will it take to generate fairness-preserving explanations?}

\begin{document}

\twocolumn[
\icmltitle{What will it take to generate fairness-preserving explanations?}



\icmlsetsymbol{equal}{*}

\begin{icmlauthorlist}
\icmlauthor{Jessica Dai}{brown}
\icmlauthor{Sohini Upadhyay}{harv}
\icmlauthor{Stephen H. Bach}{brown}
\icmlauthor{Himabindu Lakkaraju}{harv}
\end{icmlauthorlist}

\icmlaffiliation{brown}{Brown University, Providence, RI, USA}
\icmlaffiliation{harv}{Harvard University, Cambridge, MA, USA}

\icmlcorrespondingauthor{Jessica Dai}{jessica.dai@alumni.brown.edu}

\icmlkeywords{Machine Learning, ICML}

\vskip 0.3in
]



\printAffiliationsAndNotice{}  

\begin{abstract}
In situations where explanations of black-box models
may be useful, the fairness of the black-box is also often 
a relevant concern. 
However, 
the link between the fairness of the black-box model 
and the behavior of explanations for the black-box is unclear.
We focus on explanations applied to tabular datasets, suggesting that 
explanations do not necessarily preserve the fairness 
properties of the black-box algorithm. In other words, 
explanation algorithms can ignore or obscure critical relevant 
properties, creating incorrect or misleading explanations. 
More broadly, we propose future research directions for
evaluating and generating explanations such that they are
informative and relevant from a fairness perspective. 

\end{abstract}

\section{Introduction \& Motivation}

While fairness and explainability are both generally considered 
core components of ``responsible'' machine learning, surprisingly little 
work has explored the two principles in tandem. 
However, especially in light of common goals of generating 
explanations for a black-box models, it is critical that the explanation
itself can be reliably trusted to illustrate important fairness 
properties of the black-box. For example, \citet{suresh2021beyond}'s
framework for characterizing stakeholders in explainable
machine learning provides \textit{objectives} such as
debugging or improving the model,
ensuring regulatory compliance,
informing downstream actions,
justifying actions based on algorithm output,
and 
contesting a decision; and specific \textit{tasks} like 
assessing the reliability of a prediction;
detecting mistaken or discriminatory behavior;
and understanding the influence of different inputs. 
Prior work in this area has outlined similar goals 
for explanations \citep{bhatt2020machine}.
For obvious reasons, if fairness is a concern related to the 
model more broadly, it is also a critical 
consideration for these tasks and 
objectives in the context of explanations.
Furthermore, while of course calculating 
particular fairness desiderata for the underlying black-box directly
might surface unfairness, 
the stakeholders who are using an explainable ML algorithm
may not have access to the information needed for such an analysis; 
as a result, we might hope that \textit{explanations themselves} 
contain sufficient and accurate information for any
stakeholder to confidently make claims and downstream
decisions based on the explanation. 
This is especially important given that end-users
of explanations may be vulnerable to overtrusting 
or being manipulated by explanations \citep{lakkaraju2020fool}. 

However, current methods for evaluating explanations
are designed to be almost entirely application-agnostic, 
and therefore do not consider any criteria related to fairness. 
While terminology varies across the literature, commonly
used evaluation metrics for explanations include 
\textit{fidelity}, the extent to which a surrogate model
generated by an explanation algorithm produces predictions 
similar to the black-box, and 
\textit{stability}, the extent to which explanations 
generated for similar (but non-identical) inputs are similar to one another
\citep{bhatt2020evaluating, yehfidelity}.

A growing portion of the literature points to dangers in 
focusing solely on these targets when designing explanation
algorithms. \citet{slack2020much} and \citet{zhang2019should}, 
for instance, highlight
the high degree of inconsistency of explanations generated by 
perturbation-based methods under certain parameter settings---in
other words, multiple explanations generated for the same input
may result in wildly different explanations. 
\citet{kumar2020problems} and \citet{hancox2021epistemic}, meanwhile,
investigate SHAP \citep{lundberg2017unified}, and find problems 
from both technical and philosophical perspectives.
Under the framework of fairness specifically, \citet{slack2020fooling}
and \citet{aivodji2019fairwashing} illustrate specific ways 
in which either a black-box algorithm or an explanation, respectively, may be 
adversarially constructed such that the explanation, while 
having high fidelity (or achieving other desirable metrics),
misleadingly suggests that the black-box model is fair 
when in reality it is not. 
However, adversarial construction may not be necessary for 
misleading explanations to occur.

For some baseline intuition as to how fairness and explanations
may interact, consider the following. Explanations 
are often intended to provide a digestible approximation of the black-box 
algorithm's decision boundary, whether locally (in the 
neighborhood of a particular input) or globally (for all possible inputs). 
Additionally, fairness 
concerns arise when there is a meaningful difference in how
two or more demographic groups are distributed or labelled 
in the training data, 
which leads to a meaningful disparity in how the black-box
machine learning algorithm performs on the two groups 
by whatever metric one may choose \citep{corbett2018measure}.
The demographic information may or may not be used 
by the black-box. 
In the case that it is not used, 
the explanation will not \textit{explicitly} encode
information about the sensitive attribute, and an end-user
relying on the explanation alone will have little 
information about the fairness of the black-box.
In the case that it is used, such as when fairness-constrained
learning algorithms or postprocessing methods are applied,
then the black-box may learn a decision boundary such that
the boundaries are different when conditioned on 
group membership. However, explanation methods are not designed
to approximate such boundaries.
Furthermore, the explanation itself
will include information about the sensitive attribute, 
such as in the form of a feature importance score; 
it is not immediately clear what the proper interpretation
of that score should be. 
Finally, in either case, the known issue of 
isolating feature attributions
when features may be correlated with one another \citep{kumar2020problems}
is especially relevant when considering fairness applications.

\subsection{A simple example}

\begin{figure}
    \centering
    \begin{subfigure}[t]{0.22\textwidth}
        \centering
		\includegraphics[width=1\textwidth]{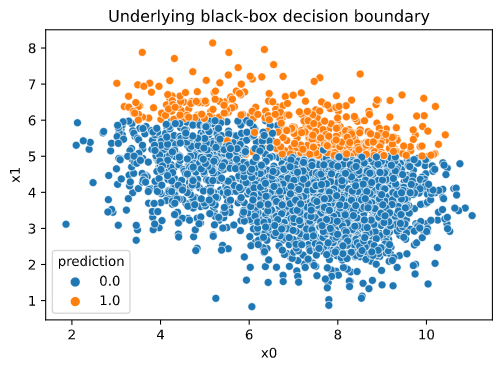}
		 \caption{Decision boundary of the black-box}
		 \label{fig:bound_bb}
    \end{subfigure}
    ~
    \begin{subfigure}[t]{0.22\textwidth}
        \centering
		\includegraphics[width=1\textwidth]{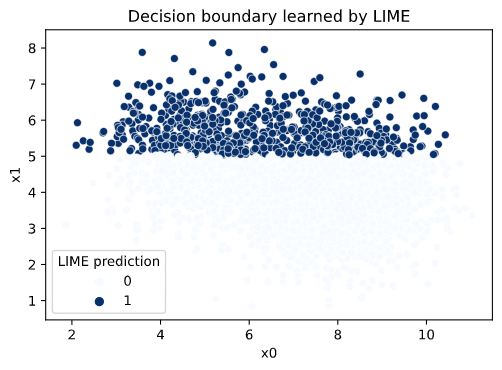}
		\caption{Decision boundary learned by LIME}
		\label{fig:bound_lime}
    \end{subfigure}
   \caption{The true decision boundaries of the black-box vs the decision boundary learned by LIME. The left cluster is the minority group, comprising 27\% of the total population.}
   \label{fig:bounds}
\end{figure}

Consider the following scenario where the black-box takes
in three features: \texttt{group}, \texttt{x0}, and \texttt{x1},
where \texttt{group} $\in \{0, 1\}$, \texttt{x0} and \texttt{x1}
are continuous, and $\texttt{x0}$ is correlated with group membership.
For some reason or another (perhaps by applying a fairness
intervention in the training process), the black-box's 
learned decision boundaries are different when conditioned on
group membership: specifically, the black-box predicts \textbf{1}
for group \textbf{0} when \texttt{x1} $> 6$, and for group \textbf{1}
when \texttt{x1} $> 5$. Figure \ref{fig:bound_bb} illustrates this black-box decision boundary.

In the case that both groups constitute $50\%$ of the population,
an explanation method that optimizes for fidelity as measured 
by performance on sampled neighbors will approximate
the decision boundary at around \texttt{x1} $ > 5.5$---an 
explanation that is 
simply incorrect for both groups
based on what we know about the black-box.
If one group is a minority of the population, however, 
the explanation's approximated decision boundary will be closer
to the majority group's decision boundary, meaning overall
better explanations for the majority group and overall
worse explanations for the minority group. 
This is illustrated in Figure \ref{fig:bound_lime}, which visualizes the decision boundary learned by LIME: note that the learned boundary is much closer to 
\texttt{x1} $>5$, the majority group's boundary, over \textit{all} data points, not
just the points corresponding to the majority group. Notably, this is a problem that seems to arise \textit{whenever} group-conditional 
decision boundaries are meaningfully distinct.
The explanations generated here, therefore, may be both
misleading and incorrect. 

\section{Our Framework}
We propose a two-part framework for further work in this area:
first, determining what constitutes a mismatch in fairness properties; 
and second, generating fairness-preserving explanations.

\subsection{Diagnosing Fairness Mismatch}

First, we provide an initial attempt at outlining what metrics
or diagnostic tests may be useful in detecting a mismatch
in fairness; these also serve, therefore, as potential criteria
or definitions for what a fairness-preserving vs fairness-obscuring
explanation may look like. These metrics are broadly motivated 
by the principle that if the model is fair, the explanations 
should not raise false alarms; similarly, if the model is unfair,
the explanations should not suggest that it is innocuous. 
In this section we attempt to pinpoint what exactly it means 
for an explanation to ``raise a false alarm'' or suggest that 
the model ``is innocuous.''

\textbf{Group fairness.} 
There are a variety of metrics through which models can be 
audited or monitored for group fairness: demographic 
parity focuses primarily on group-wise outcomes, 
while other metrics such as equalized odds, equal opportunity, 
or predictive parity, reflect some combination of 
the group-conditional confusion matrices \citep{verma2018fairness}. 

Let $\mathcal M$ represent a metric of group fairness which takes in
the predictions of some model (and potentially information 
about the true labels); $f$ represent the black-box; $E_f$ be
the surrogate model from an explanation for $f$; and 
$E_f(\vec x)$ represent evaluating the surrogate model on some 
input $\vec x$. Then, group fairness is preserved when:
$$
    | \mathcal M(f(\vec x)) - \mathcal M(E_f(\vec x)) | \leq \epsilon
$$
In other words, when, if substituting the black-box model
with the explanation's surrogate model, the predictions
generated result in similar values of the fairness metric $\mathcal M$.

Two obvious issues arise with this initial proposition. 
First, while this is straightforward for global explanation methods,
many of the most popular explanation methods like LIME
\citep{ribeiro2016should} or SHAP \citep{lundberg2017unified}
are \textit{local} explanation methods, designed to explain
specific points: that is, there is no
notion of a \textit{global} surrogate model from which group
fairness metrics can easily be calculated. 
Second, only the demographic parity metric does not require
information about the ground-truth labelling of data points;
all other metrics require this information.
The question then becomes how to determine the set of points $x$ 
on which $\mathcal M$ will be calculated for local explanations. 
One potential approach is to use the sampled points in the local 
neighborhood generated by the explanation method, and calculate
$\mathcal M$ on the neighborhood for each of the points in the 
dataset. Of course, this approach means that no ground-truth
labelling is available for this set of sampled points, and thus
the only metric that can be verified to match or mismatch
in this way is demographic parity. 

\textbf{Counterfactual fairness.}
In classification, counterfactual fairness and 
individual fairness have similar motivations: identifying how 
the prediction for a particular input $x$ would change if only
the group membership of $x$ was changed \citep{dwork2012fairness}. 
Though there is debate
about the extent to which counterfactual 
or individual fairness 
is distinct (if not orthogonal) from group 
fairness \citep{lahoti2019ifair,binns2020apparent}, 
a ``fairness-preserving'' explanation should nevertheless
capture the counterfactual behavior of the black-box model. 
To that end, let $x'$ represent the input $x$ with a changed
value for group membership, and $E_{f(x)}$ illustrate 
explanations \textit{generated for} input $x$. Then, counterfactual fairness is 
preserved when: 
$$
E_{f(x)} - E_{f(x')} \approx f(x) - f(x')
$$
In other words, when the difference between the explanation
generated for $x$ and the explanation generated for $x'$ follows
the difference between the model's behavior on $x$ and $x'$.
This abstraction also raises open questions about how exactly
the similarity should be determined. 

\textbf{Sensitive attribute.}
The treatment of the sensitive attribute in cases where it is
included in the inputs to the black-box model is also worth 
additional attention. In this case, unlike group and counterfactual
fairness, we do not propose a particular normative value
of how the sensitive attribute ought to be treated by the 
explanation algorithm in relation to the black-box.
For example, the feature importance for the
sensitive attribute being 0 does not necessarily imply that 
the black-box is not discriminatory: the influence of the sensitive
attribute may have been attributed to another, correlated 
feature. Moreover, many algorithms for fair machine learning
explicitly use the sensitive attribute in order to achieve some 
measure of fairness, such as the method proposed in \citet{hardt2016equality}.
In this sense, a feature importance of 0 might even be alarming
rather than reassuring. 
As a result, future work in this area 
may include methods which give more meaningful ways to 
interpret the influence of the sensitive attribute.

\textbf{Additional considerations.}
Finally, of note here is the distinction between \textit{evaluating
an explanation algorithm itself} for how well it preserves 
fairness properties in general, and \textit{evaluating a given, specific
explanation} for whether it is preserving relevant fairness properties
once the explanation for a particular input or model has been generated. 
These are 
different tasks---the first, for example, may be useful for a model
developer or engineer in the process of choosing an explanation method, 
while the second may be more relevant to auditing processes once
a black-box model (and a corresponding explanation algorithm) 
has been deployed. Additional work distinguishing what approaches
or metrics might be comparatively useful in either situation is warranted;
in particular, all of these proposed metrics require a comparatively
high amount of information and access to the black-box, and may
be better-suited towards the first task (evaluating algorithms
in general) rather than the second (auditing individual explanations). 

\subsection{Generating Fairness-Preserving Explanations}

As discussed above, algorithms for finding explanations typically 
focus on optimizing for metrics such as fidelity and sensitivity.
One approach for generating a fairness-preserving explanation
can be similar to early approaches to fair machine learning algorithms:
adding a penalty term in the objective function for the
extent to which the explanation is fairness-preserving 
\citep{kamishima2012fairness, zafar2017fairness}. 
For example, the original LIME objective function \citep{ribeiro2016should}
is as follows:
$$
\xi(x) = \arg\min_{g\in G} \mathcal L(f, g, \pi_x) + \Omega(g)
$$
where $\xi(x)$ is the optimal explanation for input $x$ to model $f$, 
$G$ is the class 
of sparse linear models, $\mathcal L$ is a measure of fidelity, 
$\pi_x$ is a local region around $x$,
and $\Omega$ is a measure of complexity. 
A modified objective function including a term 
such as $\psi(f, g)$ measuring the preservation of 
fairness properties described in Section 2.1, fits naturally:
$$
\xi_{\textit{fair}}(x) = \arg\min_{g\in G} \mathcal L(f, g, \pi_x) + \lambda_1\Omega(g) + \lambda_2 \psi(f, g)
$$

Here, $\lambda_1$ and $\lambda_2$ are tuning parameters for the complexity $\Omega$ and fairness-preservation term $\psi$, respectively. 

\begin{figure}
    \centering
    \includegraphics[width=0.49\textwidth]{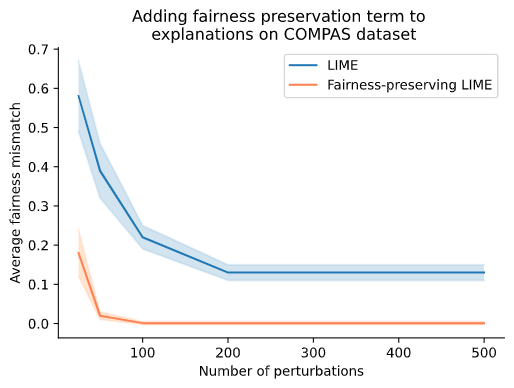}
    \caption{Original vs fairness-preserving LIME algorithms: number of perturbations
    used for LIME vs average fairness mismatch over explanations for all points in the dataset.}
    \label{fig:preserving_lime}
\end{figure}

Figure \ref{fig:preserving_lime} illustrates
the results of using this modified, fairness-preserving objective
function when finding the explanation $\xi$. 
Here, the dataset used was COMPAS; the black-box was a three-layer
deep neural net; and $\psi$ is derived from 
the group fairness equation in Section 2.1. Specifically, $\psi = |DP(f(x)) - DP(E_f(x))|$, where $DP$ is
the demographic parity metric: $P(Y=1 | S = 1) - P(Y= 1 | S = 0)$. 
In this experiment, the number of perturbations used to generate the LIME explanation was varied to show
the asymptotic fairness mismatch, as a greater number of perturbations generally 
results in a higher-certainty explanation. The fairness mismatch plotted on the y-axis is calculated exactly in the same way as $\psi$ explained above.
Our introduction of this approach is meant more as a provocation to start 
the conversation rather than a full-fledged proposal
or argument that this method is necessarily ideal; 
however, the results are promising and warrant further investigation in this direction.

\section{Discussion \& Conclusion}
In this work, we have given some intuition and preliminary results 
as to why it is important to probe the fairness of explanations: 
not just because of the often high-stakes and consequential goals 
for which explanations are used, but because
existing explanation methods focusing on metrics like fidelity may 
result in misleading and incorrect explanations 
\textit{even in the absence of an adversarial actor} constructing 
explicitly discriminatory black-boxes, 
or designing explanation methods that explicitly hide 
discrimination. Furthermore, fairness can also be viewed as a specific lens on 
performance for the model overall. In fact, the phenomenon 
illustrated in Figure \ref{fig:bounds}
can be considered to be a performance issue---strictly 
incorrect decision boundaries, though the minority group's 
decision boundary is much more 
incorrect---that can be detected by testing for fairness 
mismatch as proposed in Section 2. Of course, the 
exact behavior in this scenario may be the consequence of 
LIME's choice to focus on sparse linear models, and 
choosing a more complex interpretable model class 
(such as shallow decision trees) may alleviate the issue. 

Nevertheless, an action like this is only made possible by the first 
step of diagnosing the fairness mismatch between the black-box
and the explanation's surrogate model. 
Thinking about 
explanations in this context, therefore, also raises broader
questions about the extent to which explanations are in fact
capturing what we want; or, alternatively, ways in which the 
limitations of particular explanations or explanation methods 
may be communicated clearly to stakeholders and end-users. 

In this work, we suggested a framework for evaluating the fairness-preserving
properties of explanations,
and proposed one generic approach for producing fairness-preserving explanations.
However, this extended abstract is also meant to argue for the 
consideration of evaluation metrics for explanations more 
broadly: while fairness was the first angle we considered, there 
are undoubtedly additional necessary properties of the model---even privacy,
for example---that explanations should preserve. 

\newpage

\section*{Acknowledgements}
We would like to thank the anonymous reviewers for their insightful feedback. This work is supported in part by the NSF award \#IIS-2008461, and Google. The views expressed are those of the authors and do not reflect the official policy or position of the funding agencies.

\bibliography{papers}
\bibliographystyle{icml2021}



\end{document}